\documentclass[
twocolumn, 
hf, 
]{ceurart}

\usepackage{listings}
\usepackage{makecell}
\begin{document}

\copyrightyear{2025}
\copyrightclause{Copyright for this paper by its authors.
  Use permitted under Creative Commons License Attribution 4.0
  International (CC BY 4.0).}

\conference{\href{https://hexed-workshop.github.io}{HEXED'25: 2nd Human-Centric eXplainable AI in Education Workshop}, 20 July, 2025, Palermo, Italy}

\title{\textcolor{red}{Accepted to appear in the workshop proceedings for the HEXED'25 workshop in the 26th International Conference on Artificial Intelligence in Education 2025} \newline \newline Exploring Human–AI Complementarity in CPS Diagnosis Using Unimodal and Multimodal BERT Models}


%
\author[1]{Kester Wong}[%
orcid=0000-0002-8689-6869,
email=yew.wong.21@ucl.ac.uk
]
\cormark[1] 
\address[1]{UCL Knowledge Lab, Institute of Education, University College London, UK}

\author[2]{Sahan Bulathwela}[%
orcid=0000-0002-5878-2143,
email=m.bulathwela@ucl.ac.uk
]
\address[2]{UCL Centre for Artificial Intelligence, Department of Computer Science, University College London, UK}

\author[1]{Mutlu Cukurova}[%
orcid=0000-0001-5843-4854,
email=m.cukurova@ucl.ac.uk
]

\cortext[1]{Corresponding author.}

\begin{abstract}
  Detecting collaborative problem solving (CPS) indicators from dialogue using machine learning techniques is a significant challenge for the field of AI in Education. Recent studies have explored the use of Bidirectional Encoder Representations from Transformers (BERT) models on transcription data to reliably detect meaningful CPS indicators. A notable advancement involved the multimodal BERT variant, AudiBERT, which integrates speech and acoustic-prosodic audio features to enhance CPS diagnosis. Although initial results demonstrated multimodal improvements, the statistical significance of these enhancements remained unclear, and there was insufficient guidance on leveraging human-AI complementarity for CPS diagnosis tasks. This workshop paper extends the previous research by highlighting that the AudiBERT model not only improved the classification of classes that were sparse in the dataset, but it also had statistically significant class-wise improvements over the BERT model for classifications in the social-cognitive dimension. However, similar significant class-wise improvements over the BERT model were not observed for classifications in the affective dimension. A correlation analysis highlighted that larger training data was significantly associated with higher recall performance for both the AudiBERT and BERT models. Additionally, the precision of the BERT model was significantly associated with high inter-rater agreement among human coders. When employing the BERT model to diagnose indicators within these subskills that were well-detected by the AudiBERT model, the performance across all indicators was inconsistent. We conclude the paper by outlining a structured approach towards achieving human-AI complementarity for CPS diagnosis, highlighting the crucial inclusion of model explainability to support human agency and engagement in the reflective coding process.
\end{abstract}


\maketitle

\section{Introduction}
The use of multimodal data to perform collaborative problem solving (CPS) diagnosis has been explored in various studies. Some studies used students' multimodal interaction data to interpret the quality of CPS competence. For example, \citet{cukurova_2018NISPI} leveraged multimodal data to capture observable behaviours of students interacting and looking at resources (including human resources) around them during CPS. \citet{ouyang_2021Exploring} examined group CPS processes using multimodal data to understand the effects of different types of scaffolding on collaborative learning performance, process, and quality. These studies presented the potential of leveraging multimodal data and various analytical methods to provide insights into aspects of collaborative problem solving.

On the other hand, other studies used student interaction data to directly code for observable behaviours reflective of CPS facets, skills, or indicator descriptions. For example, \citet{stewart_2021Multimodal} explored the use of various models to automatically detect three CPS facets: construction of shared knowledge, negotiation/coordination, and maintaining team function. The study used standard and deep learning sequential classifiers (i.e., Random Forest model and Long Short-Term Memory recurrent neural networks) with various combinations of multimodal data to determine the best-performing model to diagnose these CPS facets automatically. Although the study reported that the inclusion of multimodality did not lead to improvements in model accuracy, this line of research remains inconclusive due to limited empirical work that has involved the use of models from recent AI developments (e.g., pre-trained Large Language Models) to investigate the potential of multimodality for improved CPS diagnosis.
\citet{wong_2025Rethinking} contributed to this discussion by exploring the use of a multimodal transformer-based model that processed both textual and acoustic features of speech data (i.e., AudiBERT). The study demonstrated that the inclusion of multimodal audio data could increase the range of subskills that can be accurately classified and improve the classification of negative affective states observed during CPS processes. However, there was no investigation on whether the AudiBERT model had statistically significant improvements over the BERT model. In addition, while the article highlighted the potential for leveraging multimodal models to support human CPS diagnosis tasks through human-AI complementarity, there was also no clear articulation of a suitable approach that could guide investigations on this work.

\subsection{Scope and research question}
Therefore, this study seeks to extend the work by \citet{wong_2025Rethinking} by evaluating the statistical significance in improvements observed between the AudiBERT and BERT models for detecting CPS subskill and affective state classes. Based on this analysis, we examine factors that could be associated with model performance and investigate the classification of indicators within well-detected subskill and affective state classes. The aim is to conceptualise a structured approach towards human-AI complementarity using multimodal and unimodal models for CPS diagnosis at the indicator level. The following are the three research questions posed in this study.
\begin{enumerate}
    \item[RQ1:] Are the class-wise differences in model performance between the multimodal AudiBERT model and the unimodal BERT model statistically significant in the classification of CPS subskill and affective state classes?
    
    \item [RQ2:] How do the scarcity of data and the complexity of CPS subskill and affective state classes affect the classification performance of the both models?
    
    \item [RQ3:] Do all fine-grained indicators within well-detected CPS subskill and affective state classes demonstrate similarly high classification performance using the BERT model?
    
\end{enumerate}

\section{Method}
The dataset used in this study consists of transcription and acoustic-prosodic features extracted from audio data at the utterance level. These utterances were coded at the indicator level, with each indicator corresponding to one of the 10 CPS subskills in the social-cognitive dimension; eight problem solving subskills (i.e., SS1 - SS8), and two scripting subskills (i.e., SC1 \& SC2). If suitable, the utterance was also coded concurrently with an indicator corresponding to one of three affective states (i.e., AS1 - AS3) in the affective dimension (see \cite{wong_2025Rethinking} for CPS framework). The dataset was obtained from 78 Secondary school students (aged 14 - 15 years) from a public school. Participants worked in triads to solve a mathematics problem solving question in a given CPS task on a remote video conferencing platform without teacher intervention.

\subsection{Analysis of differences in model performance}
The analysis was performed at the subskill level for the social-cognitive dimension ($n = 10$ classes) and separately for the affective dimension ($n = 3$ classes). The class-wise differences in performance between the AudiBERT model and the BERT model for each dimension were investigated by conducting a non-parametric one-tailed Wilcoxon signed-rank test, since the number of classes involved in the comparison was small ($n < 30$). The effect size was approximated using the rank-biserial correlation \cite{minium_1993Statistical}, with the statistical power calculation being the same as that for a paired t-test with an adjustment to the sample size \cite{guenther_1982Normal}.

\subsection{Analysis of class complexity and data scarcity}
Correlation analysis was performed to understand how the complexity of CPS labels and data scarcity were associated with the BERT model's performance and the AudiBERT model's performance, respectively, in classifying the CPS subskill and affective state classes ($n = 13$ classes). In particular, the analysis involved correlating the weighted F1, precision, and recall scores of each class with Cohen's kappa of human labelling agreements (which was used as a proxy to quantify the complexity of coding CPS), and separately with the number of coded utterances in the training dataset for the respective classes (which was used as a proxy to quantify data scarcity in the dataset). Spearman and Pearson correlation analyses were used to obtain a comprehensive understanding of the relationships between these variables. Pearson's approach assumes the linearity of the relationship, while Spearman's non-parametric approach assumes monotonic trends that may not follow a linear trend. Bonferroni correction was not applied since such a correction would be overly conservative for the  analyses undertaken, and obscure meaningful results due to inflated type II error risk \cite{armstrong_2014When}. Spearman's and Pearson's correlation coefficients are respectively annotated as $\rho$ and $r$ in this paper.

However, Cohen's kappa could also be associated with the number of samples that coders are applying the coding to, as larger sample sizes may lead to more stable and reliable agreements. Hence, prior to this analysis, the correlation between Cohen's kappa and testing data was examined to determine whether variability in kappa values might be associated with class size, and confound the correlation analysis between Cohen's kappa and model performance. It was found that the correlation between Cohen's kappa and testing data was weak and not statistically significant ($\rho = 0.121, p = .708; r = -0.147, p = .649$).

\subsection {Classification of indicators within well-detected classes}
The following subskill and affective state classes were considered well-detected by \citet{wong_2025Rethinking}, in which their classifications by the BERT model and the AudiBERT model both had a weighted F1 score of above 0.60: `\textit{Building shared understanding}' (SS2), `\textit{Using scripting}' (SC1), `\textit{Regulating scripting}' (SC2), `\textit{Neutral affective state}' (AS1), `\textit{Negative affective state}' (AS2), `\textit{Positive affective state}' (AS3). Since the affective states AS1 and AS3 each involve one indicator from the dataset, the classification of fine-grained indicators was performed separately for SS2 (11 indicators), for SC1 and SC2 (5 indicators), and for AS2 (6 indicators). The implementation of the BERT model follows the method used by \citet{wong_2025Rethinking}. Table \ref{tab:subskilltable} provides a breakdown of the training and testing dataset of the indicators involved in the classification.


\begin{table*}
\caption{Dataset distribution of indicators across `\textit{Building shared understanding}' problem solving subskill (SS2), scripting subskill (SC1 \& SC2), and negative affective state (AS2).}
\label{tab:subskilltable}
  \resizebox{\textwidth}{!}{
\begin{tabular}{c|ccccccccccc|ccccc|cccccc}
\hline
Category & \multicolumn{11}{c|}{`\textit{Building shared understanding}' problem solving subskill} & \multicolumn{5}{c|}{Scripting subskill} & \multicolumn{6}{c}{Negative affective state} \\
Indicator & PS04 & PS05 & PS06 & PS07 & PS08 & PS09 & PS11 & PS12 & PS13 & PS15 & PS16 & S01 & S02 & S03 & S04 & S05 & A02 & A03 & A04 & A05 & A07 & A08 \\
\hline
Training & 316 & 148 & 99  & 34  & 31  & 14  & 39  & 48  & 8   & 112 & 126 & 426 & 65  & 781 & 677 & 10  & 328 & 94  & 246 & 45  & 10  & 10  \\
Testing & 79  & 37  & 25  & 8   & 8   & 3   & 10  & 12  & 2   & 28  & 32  & 107 & 16  & 195 & 169 & 3   & 82  & 23  & 62  & 11  & 3   & 3   \\
\hline
\end{tabular}
}
\end{table*}

Classification of the indicators was performed using the BERT model instead of the AudiBERT model. This is because there are available methods to enable interpretability of the BERT model \cite{Lundberg2017}, yet there is currently no architecture to allow for such explanations of the AudiBERT model. In proposing an approach to human-AI complementarity for CPS diagnosis tasks, incorporating model explainability can support researchers and practitioners in exercising their agency and engaging in the reflective coding process.

\section{Results}
\subsection{RQ1: Significance test for class-wise differences in model performance}
The class-wise comparison conducted using the one-tailed Wilcoxon signed-ranked test showed that the AudiBERT model significantly outperformed the BERT model for classifications of the CPS subskill classes ($W = 20.0, p = .031 < .05$), with an approximate effect size of 0.91 and statistical power of 0.84. However, this was not observed for classifications of the affective state class, as the AudiBERT model did not have significant improvements over the BERT model ($W = 3.0, p = .625$), with an approximate effect size of 0 and statistical power of 0.05. Yet, it was observed by \citet{wong_2025Rethinking} that the AudiBERT was able to improve the detection of classes that were sparse in the dataset to improve the classification of the positive affective state. Hence, it would be more beneficial to use the multimodal AudiBERT model for diagnosing CPS subskills and affective states, rather than the unimodal BERT model. 

\subsection{RQ2: Impact of data scarcity and label complexity on model performance}

\subsubsection{Data scarcity}
The correlation between the precision scores and the number of training data was weak and not statistically significant for both the AudiBERT model ($\rho = 0.098, p = .7621; r = 0.334, p = .2889$) and the BERT model ($\rho = -0.025, p = .9397; r = 0.217, p = .4980$). However, the correlation between the recall scores and the number of training data was much higher and statistically significant for both the AudiBERT model ($\rho = 0.622, p = .031 < .05; r = 0.637, p = .026 < .05$) and the BERT model ($\rho = 0.748, p = .0051 < .01; r = 0.772, p = .0033 < .01$). For the F1 score, a smaller correlation with the number of training data was found, with only the Pearson correlation having statistical significance for the AudiBERT model ($\rho = 0.517, p = .09; r = 0.596, p = .041 < .05$). However, the correlation between F1 score and the number of training data was statistically significant for the BERT model ($\rho = 0.708, p = .0101 < .05; r = 0.703, p = .0108 < .05$). These findings suggest that both the AudiBERT and BERT models are effective in identifying instances of a positive class when more training examples are provided. As expected, due to their transformer architectures with attention heads, the performance of both the AudiBERT and BERT models improves as the amount of training data increases. Table \ref{tab:datacorrelation_metrics} summarises the correlation analysis of the two models (i.e., the AudiBERT model and the BERT model) respectively to the number of training data for the subskills and affective states. 

\begin{table}[ht]
\caption{Correlation between classification performance and the number of training data. The higher correlation value between the AudiBERT model and the BERT model for each performance metric is indicated in bold face, with $*: p < 0.05$, $**: p < 0.01$.}
\label{tab:datacorrelation_metrics}
\centering
  \resizebox{0.47\textwidth}{!}{
\begin{tabular}{lcc}
\hline
Model & \makecell{Spearman Correlation \\ $\boldsymbol{\rho}$ (p-value)} & \makecell{Pearson Correlation \\$r$ (p-value)} \\
\hline
\multicolumn{3}{c}{Precision} \\
\hline
BERT        & $-0.025$ (.9397)             & 0.217 (.4980) \\
AudiBERT    & \textbf{0.098} (.7621)       & \textbf{0.334} (.2889) \\
\hline
\multicolumn{3}{c}{Recall} \\
\hline
BERT        & \textbf{0.748**} (.0051)     & \textbf{0.772**} (.0033) \\
AudiBERT    & 0.622* (.031)                & 0.637* (.026) \\
\hline
\multicolumn{3}{c}{F1 Score} \\
\hline
BERT        & \textbf{0.708*} (.0101)      & \textbf{0.703*} (.0108) \\
AudiBERT    & 0.517 (.09)                  & 0.596* (.041) \\
\hline
\end{tabular}
}
\end{table}

\subsubsection{Task complexity of coding CPS indicators}
Table \ref{tab:correlation_metrics} presents the correlation analysis of the two models (i.e., the AudiBERT model and the BERT model) respectively to Cohen's kappa across the coded subskills and affective states. 

\begin{table}[ht]
\caption{Correlation between classification performance and Cohen's kappa. The higher correlation value between the AudiBERT model and the BERT model for each performance metric is indicated in bold face, with $*: p < 0.05$.
}
\label{tab:correlation_metrics}
\centering
  \resizebox{0.47\textwidth}{!}{
\begin{tabular}{lcc}
\hline
Model & \makecell{Spearman Correlation \\ $\boldsymbol{\rho}$ (p-value)} & \makecell{Pearson Correlation \\$r$ (p-value)} \\
\hline
\multicolumn{3}{c}{Precision} \\
\hline
BERT   & \textbf{0.602* (.0382)} & 0.508 (.0916) \\
AudiBERT  & 0.497 (.1006)    & 0.\textbf{517 (.0852)} \\
\hline
\multicolumn{3}{c}{Recall} \\
\hline
BERT            & 0.245 (.4433)     & 0.320 (.3102) \\
AudiBERT  & \textbf{0.399 (.1993)}     & \textbf{0.421 (.1728)} \\
\hline
\multicolumn{3}{c}{F1} \\
\hline
BERT            & 0.319 (.3126)     & 0.402 (.1951) \\
AudiBERT  & \textbf{0.455 (.1377)  }   & \textbf{0.454 (.1384)} \\
\hline
\end{tabular}
}

\end{table}

It is observed that only the Spearman correlation between the precision performance of the BERT model and Cohen's kappa was statistically significant ($\rho = 0.602, p = .0382 < .05$). This implies that correct positive classifications on classes by the BERT model had a mild correlation with classes that humans also found to be clearer and less complex. However, since the Pearson correlation was not significant ($r = 0.508, p = .0916$), it suggests that the correlation relationship was not strictly linear. The other correlations did not present any statistical significance. However, it should also be noted that the lack of statistical significance may be attributed to the limited number of subskills and affective states included in the correlation analysis. 

\subsection{RQ3: Classification of indicators within well-detected subskills and affective state}
Table \ref{tab:performance_table} provides an overview of the BERT model's performance for the classification of indicators in the well-detected classes. It is observed that the weighted F1 score for the classification of indicators in the negative affective states outperforms the classification of indicators in the scripting subskill. The performance for the classification of indicators in the ‘\textit{Building shared understanding}’ problem solving subskill was the lowest among the three classifications across all the metrics. This could be due to SS2 having almost double the number of indicators compared to the others. 

\begin{table}[!h]
\caption{BERT model performance for the classification of indicators in the `\textit{Building shared understanding}'  problem solving subskill (SS2), the scripting subskills (SC1 \& SC2) and the negative affective state (AS2) using accuracy, precision, recall, F1 score.}
\label{tab:performance_table}
\small
\setlength{\tabcolsep}{10pt}
\centering
\resizebox{0.47\textwidth}{!}{
\begin{tabular}{l|cccc}
\hline
Category & Accuracy & Precision & Recall & F1 score \\ 
\hline
SS2 & 0.480 & 0.462 & 0.480 & 0.411 \\
SC1 \& SC2 & 0.663 & 0.642 & 0.663 & 0.649 \\ 
AS2 & 0.804 & 0.794 & 0.804 & 0.792 \\
\hline
\end{tabular}
}
\end{table}

Table \ref{tab:combined} shows the performance of the classified indicators within the respective subskills and affective state by the BERT model. It is observed that only the `\textit{Asking questions to clarify understanding, ideas or contributions}' (PS04), `\textit{Responding to script components}' (S03), `\textit{Discussing work status and progress on script components}' (S04), `\textit{Verbal cues to express bewilderment from observations in the task or of others}' (A02), `\textit{Use of strong and vulgar expressions}' (A03), and `\textit{Verbalising difficulty or dislike of work to others}' (A04) were well-detected indicators (i.e., $\text{weighted F1 score} \geq 0.60$). Hence, it is clear that not all fine-grained indicators within well-detected CPS subskills and affective states could be detected with similar classification performance by the BERT model. 

\begin{table}[!h]
\caption{BERT model classification report of indicators with metrics as columns, grouped by subskills and affective state. The best and second-best performances are indicated in \textbf{bold} and \emph{italic} faces, respectively.}
\label{tab:combined}
\small
\setlength{\tabcolsep}{10pt}
\centering
\resizebox{0.47\textwidth}{!}{
\begin{tabular}{l|cccc}
\hline
Indicator & Precision & Recall & F1 score & Support \\
\hline
\multicolumn{5}{c}{`\textit{Building shared understanding}' CPS subskill} \\
\hline
PS04 & \textbf{0.655} & \textbf{0.911} & \textbf{0.762} & 79 \\
PS05 & 0.263 & 0.541 & 0.354 & 37 \\
PS06 & 0.444 & 0.160 & 0.235 & 25 \\
PS07 & 0.000 & 0.000 & 0.000 & 8 \\
PS08 & 0.000 & 0.000 & 0.000 & 8 \\
PS09 & 0.000 & 0.000 & 0.000 & 3 \\
PS11 & 0.000 & 0.000 & 0.000 & 10 \\
PS12 & 0.000 & 0.000 & 0.000 & 12 \\
PS13 & 0.000 & 0.000 & 0.000 & 2 \\
PS15 & 1.000 & 0.143 & 0.250 & 28 \\
PS16 & \textit{0.378} & \textit{0.531} & \textit{0.442} & 32 \\
\hline
\multicolumn{5}{c}{Scripting subskill} \\
\midrule
S01 & 0.523 & 0.533 & 0.528 & 107 \\
S02 & 0.000 & 0.000 & 0.000 & 16 \\
S03 & \textit{0.668} & \textbf{0.795} & \textbf{0.726} & 195 \\
S04 & \textbf{0.758} & \textit{0.669} & \textit{0.711} & 169 \\
S05 & 0.000 & 0.000 & 0.000 & 3 \\
\hline
\multicolumn{5}{c}{Negative affective state} \\
\hline
A02 & \textbf{0.904} & \textit{0.805} & \textit{0.852} & 82 \\
A03 & 0.545 & 0.783 & 0.643 & 23 \\
A04 & \textit{0.831} & \textbf{0.952} & \textbf{0.887} & 62 \\
A05 & 0.714 & 0.455 & 0.556 & 11 \\
A07 & 0.000 & 0.000 & 0.000 & 3 \\
A08 & 0.000 & 0.000 & 0.000 & 3 \\
\hline
\end{tabular}
}

\end{table}





\section{Discussion}
\citet{cukurova2025interplay} articulates that ``\textit{...the current complementarity paradigm is to make a better match of what humans can do and what AI can do with the problems to be tackled to achieve productivity gains at tasks rather than making humans more intelligent.}" In this paper, human-AI complementarity requires the active engagement of human coders through the usage of AI to achieve more consistent and reliable coding of CPS indicators. Hence, based on the results observed, a structured approach is proposed to leverage human-AI complementarity for CPS diagnosis at the indicator level. 

Prior to any automation, human coders would perform coding on a subset of the dataset (e.g., 10\% - 15\%) to obtain inter-rater reliability using Cohen's kappa. Since the AudiBERT model was able to significantly outperform the BERT model in classifying the subskill and affective state classes, it could be used to provide the initial classification across the given dataset. Then, the human coder would be presented with the classification results and asked to select an indicator out of the list of indicators that is associated with the corresponding subskill or affective state class.

In order to guide the human coder in selecting an indicator, at least three pieces of information should be provided. Firstly, a BERT model would be used to obtain and provide the classification of the indicator within a well-detected subskill or affective state class. Secondly, information on the tokenised words in the BERT model that drive the classification would also be provided. Thirdly, a warning to consider whether the classifications are suitable (i.e., the subskills and affective state class, and the indicator class) would be prompted to human coders. This warning is provided when the class suggested to the human coder was trained on a dataset size smaller than the median dataset size across all the classes. The human coder can then make an informed choice to accept the suggested indicator, or make amendments to the choice of indicators.

However, if the human coder is unable to determine a suitable indicator in the suggested subskills and affective state class, an alternative subskills and affective state class should be provided to the human coder using the BERT model. Since the precision of the BERT model was significantly correlated with Cohen's kappa, this suggests that the BERT model could be used to automate classes that achieved high consensus during the inter-rater reliability test. Based on the inter-rater reliability of the dataset subset, subskill and affective state classes with high Cohen's kappa values (e.g., $\geq 0.80$) would indicate that it is of lesser complexity, and the BERT model can be used to automate the second suggestion for these subskill and affective state classes. If the two suggestions are the same, the human coder will have to decide from the list of all the possible indicators to determine the appropriate coding for the particular utterance.

Since the AudiBERT model outperforms the BERT model in detecting the subskill and affective state classes, it would be more suitable to explore the use of the AudiBERT model to classify indicators. However, the impetus for proposing the use of a BERT model to perform classification of indicators stems from the current lack of development in applying Explainable AI (XAI) techniques such as SHAP or LIME to determine the tokenised words and the audio features that are contributing to the classifications by the AudiBERT model. Although there are studies that have sought to use XAI techniques to explain audio analysis \cite{BECKER2024418}, the AudiBERT model involves the concatenation of a text embedding from the BERT model and an audio embedding from Wav2Vec2.0 that were each obtained from a complex architecture that involves a Bidirectional Long-Short Term Memory (BiLSTM) network and a self-attention layer \cite{toto_2021AudiBERT}. Therefore, in view of proposing an approach that enables human-AI complementarity, importance is placed more on existing available approaches that support human coders in the coding process, rather than simply using a model with potentially better performance.

\section{Future work}
It is important to highlight that the proposed human-AI complementarity scenario for CPS diagnosis is rather hypothetical at this stage and requires rigorous empirical investigations and refinement. Regardless, it highlights an approach that leverages the strengths of a multimodal model (i.e., AudiBERT) for improved classification, as well as the strengths of an unimodal model (i.e., BERT) for providing model explainability, thereby enabling human-AI complementarity for CPS diagnosis. Further work can be undertaken to design an ensemble architecture that supports this approach and test it with human coders to investigate its performance and usage. The ensemble model can subsequently be applied in pilot studies involving human-in-the-loop implementations to evaluate the conceptualised complementarity workflow with users such as educators and learning science researchers. Such studies should also investigate the pedagogical implications of possible misclassification on learners and teachers.

Additionally, the design of an architecture that can enable model explanation of the AudiBERT model should be further explored. This could involve examining the time frame for the audio features that contribute to the AudiBERT classification and determining whether the tokenised words that were used also contribute to the AudiBERT classification within the same time frame. The development of such an architecture opens up numerous possible research investigations, such as understanding why the multimodal AudiBERT model outperforms the BERT model, and what audio features and tokenised words contribute to the accurate classification of CPS subskills and affective states. The provision of such information can support users of such models (e.g., researchers and educators) to reflect upon the provided information and guide their decisions relating to CPS diagnosis through human-AI feedback loops \cite{glickman_2024How,holstein_2023Supporting}.

\begin{acknowledgments}
\small 
This work is funded by the European Commission's projects ``Teacher-AI Complementarity (TaiCo)" (Project ID: 101177268), ``Humane AI" (Grant No. 820437) and ``X5GON" (Grant No. 761758). The CPS framework and classification report are provided at \url{https://osf.io/u4nvb/?view\_only=3b02f769ea4746a0bbbae514ab1acb8e}.
\end{acknowledgments}

\section*{Declaration on Generative AI}
  The authors have not employed any Generative AI tools.



\begin{thebibliography}{13}
\expandafter\ifx\csname natexlab\endcsname\relax\def\natexlab#1{#1}\fi
\providecommand{\url}[1]{\texttt{#1}}
\providecommand{\href}[2]{#2}
\providecommand{\path}[1]{#1}
\providecommand{\DOIprefix}{doi:}
\providecommand{\ArXivprefix}{arXiv:}
\providecommand{\URLprefix}{URL: }
\providecommand{\Pubmedprefix}{pmid:}
\providecommand{\doi}[1]{\href{http://dx.doi.org/#1}{\path{#1}}}
\providecommand{\Pubmed}[1]{\href{pmid:#1}{\path{#1}}}
\providecommand{\bibinfo}[2]{#2}
\ifx\xfnm\relax \def\xfnm[#1]{\unskip,\space#1}\fi
\bibitem[{Cukurova et~al.(2018)Cukurova, Luckin, Mill{\'a}n, and Mavrikis}]{cukurova_2018NISPI}
\bibinfo{author}{M.~Cukurova}, \bibinfo{author}{R.~Luckin}, \bibinfo{author}{E.~Mill{\'a}n}, \bibinfo{author}{M.~Mavrikis},
\newblock \bibinfo{title}{The {{NISPI Framework}}: {{Analysing Collaborative Problem-Solving}} from {{Students}}' {{Physical Interactions}}},
\newblock \bibinfo{journal}{Computers \& Education} \bibinfo{volume}{116} (\bibinfo{year}{2018}) \bibinfo{pages}{93--109}. \DOIprefix\doi{10.1016/j.compedu.2017.08.007}.
\bibitem[{Ouyang et~al.(2021)Ouyang, Chen, Cheng, Tang, and Su}]{ouyang_2021Exploring}
\bibinfo{author}{F.~Ouyang}, \bibinfo{author}{Z.~Chen}, \bibinfo{author}{M.~Cheng}, \bibinfo{author}{Z.~Tang}, \bibinfo{author}{C.-Y. Su},
\newblock \bibinfo{title}{Exploring the effect of three scaffoldings on the collaborative problem-solving processes in {{China}}'s higher education},
\newblock \bibinfo{journal}{International Journal of Educational Technology in Higher Education} \bibinfo{volume}{18} (\bibinfo{year}{2021}) \bibinfo{pages}{35}. \DOIprefix\doi{10.1186/s41239-021-00273-y}.
\bibitem[{Stewart et~al.(2021)Stewart, Keirn, and D'Mello}]{stewart_2021Multimodal}
\bibinfo{author}{A.~E.~B. Stewart}, \bibinfo{author}{Z.~Keirn}, \bibinfo{author}{S.~K. D'Mello},
\newblock \bibinfo{title}{Multimodal modeling of collaborative problem-solving facets in triads},
\newblock \bibinfo{journal}{User Modeling and User-Adapted Interaction} \bibinfo{volume}{31} (\bibinfo{year}{2021}) \bibinfo{pages}{713--751}. \DOIprefix\doi{10.1007/s11257-021-09290-y}.
\bibitem[{Wong et~al.(2025)Wong, Wu, Bulathwela, and Cukurova}]{wong_2025Rethinking}
\bibinfo{author}{K.~Wong}, \bibinfo{author}{B.~Wu}, \bibinfo{author}{S.~Bulathwela}, \bibinfo{author}{M.~Cukurova},
\newblock \bibinfo{title}{Rethinking the potential of multimodality in collaborative problem solving diagnosis with large language models},
\newblock in: \bibinfo{editor}{A.~I. Cristea}, \bibinfo{editor}{E.~Walker}, \bibinfo{editor}{Y.~Lu}, \bibinfo{editor}{O.~C. Santos}, \bibinfo{editor}{S.~Isotani} (Eds.), \bibinfo{booktitle}{Artificial Intelligence in Education}, \bibinfo{publisher}{Springer Nature Switzerland}, \bibinfo{address}{Cham}, \bibinfo{year}{2025}, pp. \bibinfo{pages}{18--32}. \DOIprefix\doi{10.1007/978-3-031-98417-4_2}.
\bibitem[{Minium et~al.(1993)Minium, King, and Bear}]{minium_1993Statistical}
\bibinfo{author}{E.~W. Minium}, \bibinfo{author}{B.~M. King}, \bibinfo{author}{G.~R. Bear}, \bibinfo{title}{Statistical Reasoning in Psychology and Education}, \bibinfo{edition}{3rd ed} ed., \bibinfo{publisher}{Wiley}, \bibinfo{address}{New York}, \bibinfo{year}{1993}.
\bibitem[{Guenther(1982)}]{guenther_1982Normal}
\bibinfo{author}{W.~C. Guenther},
\newblock \bibinfo{title}{Normal theory sample size formulas for some nonnormal distributions},
\newblock \bibinfo{journal}{Communications in Statistics - Simulation and Computation} \bibinfo{volume}{11} (\bibinfo{year}{1982}) \bibinfo{pages}{727--732}. \DOIprefix\doi{10.1080/03610918208812291}.
\bibitem[{Armstrong(2014)}]{armstrong_2014When}
\bibinfo{author}{R.~A. Armstrong},
\newblock \bibinfo{title}{When to use the {{Bonferroni}} correction},
\newblock \bibinfo{journal}{Ophthalmic and Physiological Optics} \bibinfo{volume}{34} (\bibinfo{year}{2014}) \bibinfo{pages}{502--508}. \DOIprefix\doi{10.1111/opo.12131}.
\bibitem[{Lundberg and Lee(2017)}]{Lundberg2017}
\bibinfo{author}{S.~M. Lundberg}, \bibinfo{author}{S.-I. Lee},
\newblock \bibinfo{title}{A unified approach to interpreting model predictions},
\newblock in: \bibinfo{booktitle}{Proceedings of the 31st International Conference on Neural Information Processing Systems}, NIPS'17, \bibinfo{publisher}{Curran Associates Inc.}, \bibinfo{address}{Red Hook, NY, USA}, \bibinfo{year}{2017}, p. \bibinfo{pages}{4768–4777}.
\bibitem[{Cukurova(2025)}]{cukurova2025interplay}
\bibinfo{author}{M.~Cukurova},
\newblock \bibinfo{title}{The interplay of learning, analytics and artificial intelligence in education: A vision for hybrid intelligence},
\newblock \bibinfo{journal}{British Journal of Educational Technology} \bibinfo{volume}{56} (\bibinfo{year}{2025}) \bibinfo{pages}{469--488}.
\bibitem[{Becker et~al.(2024)Becker, Vielhaben, Ackermann, Müller, Lapuschkin, and Samek}]{BECKER2024418}
\bibinfo{author}{S.~Becker}, \bibinfo{author}{J.~Vielhaben}, \bibinfo{author}{M.~Ackermann}, \bibinfo{author}{K.-R. Müller}, \bibinfo{author}{S.~Lapuschkin}, \bibinfo{author}{W.~Samek},
\newblock \bibinfo{title}{Audiomnist: Exploring explainable artificial intelligence for audio analysis on a simple benchmark},
\newblock \bibinfo{journal}{Journal of the Franklin Institute} \bibinfo{volume}{361} (\bibinfo{year}{2024}) \bibinfo{pages}{418--428}. \DOIprefix\doi{https://doi.org/10.1016/j.jfranklin.2023.11.038}.
\bibitem[{Toto et~al.(2021)Toto, Tlachac, and Rundensteiner}]{toto_2021AudiBERT}
\bibinfo{author}{E.~Toto}, \bibinfo{author}{M.~Tlachac}, \bibinfo{author}{E.~A. Rundensteiner},
\newblock \bibinfo{title}{{{AudiBERT}}: {{A Deep Transfer Learning Multimodal Classification Framework}} for {{Depression Screening}}},
\newblock in: \bibinfo{booktitle}{Proceedings of the 30th {{ACM International Conference}} on {{Information}} \& {{Knowledge Management}}}, \bibinfo{publisher}{ACM}, \bibinfo{year}{2021}, pp. \bibinfo{pages}{4145--4154}. \DOIprefix\doi{10.1145/3459637.3481895}.
\bibitem[{Glickman and Sharot(2024)}]{glickman_2024How}
\bibinfo{author}{M.~Glickman}, \bibinfo{author}{T.~Sharot},
\newblock \bibinfo{title}{How human--{{AI}} feedback loops alter human perceptual, emotional and social judgements},
\newblock \bibinfo{journal}{Nature Human Behaviour} \bibinfo{volume}{9} (\bibinfo{year}{2024}) \bibinfo{pages}{345--359}. \DOIprefix\doi{10.1038/s41562-024-02077-2}.
\bibitem[{Holstein et~al.(2023)Holstein, {De-Arteaga}, Tumati, and Cheng}]{holstein_2023Supporting}
\bibinfo{author}{K.~Holstein}, \bibinfo{author}{M.~{De-Arteaga}}, \bibinfo{author}{L.~Tumati}, \bibinfo{author}{Y.~Cheng},
\newblock \bibinfo{title}{Toward {{Supporting Perceptual Complementarity}} in {{Human-AI Collaboration}} via {{Reflection}} on {{Unobservables}}},
\newblock \bibinfo{journal}{Proceedings of the ACM on Human-Computer Interaction} \bibinfo{volume}{7} (\bibinfo{year}{2023}) \bibinfo{pages}{1--20}. \DOIprefix\doi{10.1145/3579628}.

\end{thebibliography}
\end{document}